\pdfoutput=1

\documentclass[11pt]{article}

\usepackage[hyperref]{emnlp2021}

\usepackage{times}
\usepackage{latexsym}

\usepackage[T1]{fontenc}

\usepackage[utf8]{inputenc}

\usepackage{microtype}

\usepackage{xcolor}

\usepackage{booktabs}
\usepackage{multirow}
\usepackage{graphicx}
\usepackage{bm}
\usepackage{makecell}
\usepackage{amsmath}
\usepackage[ruled]{algorithm2e}

\usepackage{etoolbox}
\makeatletter
\patchcmd{\@algocf@start}
  {-1.5em}
  {0pt}
  {}{}
\makeatother

\newenvironment{citemize}{\begin{list}{$\bullet$}{\topsep=.2\smallskipamount\itemsep=0pt\parsep=1pt\labelwidth=.5em}}{\end{list}}
\newenvironment{cenumerate}{\begin{list}{\labelenumi}{\usecounter{enumi}\topsep=.2\smallskipamount\itemsep=0pt\parsep=1pt\labelwidth=1em}}{\end{list}}

\title{Subword-level Transformations for Non-Autoregressive GEC Tagging}
\title{Non-autoregressive Subword-level Tagging for Grammatical Error Correction}
\title{Subword-level Transformations for GEC-Tagging Approach}
\title{Subword-level Transformations for GEC Tagging}
\title{Constructing Subword-level Transformations for GEC Tagging}
\title{On Constructing Subword-level Transformations for GEC-Tagging Approach}
\title{Character Transformations for Non-Autoregressive GEC Tagging}

\author{
  Milan Straka \and Jakub Náplava \and Jana Straková\\
  Charles University, \\
  Faculty of Mathematics and Physics, \\
  Institute of Formal and Applied Linguistics \\
  \texttt{\{straka,naplava,strakova\}@ufal.mff.cuni.cz}
}

\usepackage[absolute]{textpos}
\begin{document}
\begin{textblock}{16}(0,0.1)\centerline{This paper was published in \textbf{W-NUT 2021} -- please cite the published version {\small\url{https://aclanthology.org/2021.wnut-1.46}}.}\end{textblock}
\maketitle
\begin{abstract}

We propose a character-based non-autoregressive GEC approach, with automatically generated character transformations. Recently, per-word classification of correction edits has proven an efficient, parallelizable alternative to current encoder-decoder GEC systems. We show that word replacement edits may be suboptimal and lead to explosion of rules for spelling, diacritization and errors in morphologically rich languages, and propose a method for generating character transformations from GEC corpus. Finally, we train character transformation models for Czech, German and Russian, reaching solid results and dramatic speedup compared to autoregressive systems. The source code is released at\penalty-100{\small\url{https://github.com/ufal/wnut2021_character_transformations_gec}}.

\end{abstract}

\section{Introduction}

The current state of the art for grammatical error correction (GEC) is achieved with  encoder-decoder architectures, leveraging large models with enormous computational demands \cite{grundkiewicz-etal-2019-neural,Rothe-2021-mT5-GEC-SOTA}. As such autoregressive approach is slow on inference and is impossible to parallelize, it has recently been suggested to replace autoregressive sequence-to-sequence decoding with per-token tagging to enable parallel decoding, achieving a dramatic speedup by a factor of 10 in NMT \cite{gu2018nonautoregressive} and very recently, also in GEC \cite{omelianchuk-etal-2020-gector}.

\citet{omelianchuk-etal-2020-gector} approaches GEC as a tagging task, discriminating between a set of word-level transformations. The designed set is efficient for English corpora, which rarely contain spelling errors, and for English language, which does not have diacritization marks and its morphology is very modest compared to morphologically rich languages such as Czech or Russian. Using a set of word-level transformations designed for English, all character-level corrections would have to be handled by generic word-for-word REPLACE rule, leading to an explosion of rules.

We therefore suggest character transformations on subword level. Moreover, our transformations are automatically inferred from the corpus as opposed to being manually designed. Our approach has the following advantages:

\begin{citemize}
    \item character-level errors, such as diacritics, spelling and morphology are handled,
    \item the transformations can be shared between subwords, preventing an explosion of rules,
    \item the transformations are generated automatically from corpus for each language.
\end{citemize}

We present an oracle analysis of various transformations sets at different levels, in English and three other languages: Czech, German and Russian. We find that the word-level set of rules may be suboptimal for morphologically rich languages and corpora with spelling errors and diacritics.

Finally, we train models with character transformations for non-autoregressive grammatical error correction in Czech, German and Russian, reaching solid results and dramatic speedup compared to autoregressive systems.

\section{Related Work}


\citet{awasthi-etal-2019-parallel} propose an alternative to popular encoder-decoder architecture for GEC: a sequence-to-edit model which labels words with edits. Its advantage is parallel decoding while keeping competitive results. \citet{mallinson-etal-2020-felix} introduce a framework consisting of two tasks: \textit{tagging}, which chooses and arbitrarily reorders a subset of input tokens to keep, and \textit{insertion}, which in-fills the missing tokens with another pretrained masked language model. \citet{omelianchuk-etal-2020-gector} develop custom, manually designed, per-token \textit{g-transformations}.

We further improve the sequence-to-edit model with an attempt at non-autoregressive grammatical error correction for languages other than English, with \textit{character} transformations applied at \textit{subwords}, inferred \textit{automatically} from parallel GEC corpus.

\section{Transformations}
\label{sec:transformations}

Given that we encode an input sentence using BERT \cite{devlin-etal-2019-bert}, it is natural to represent it using a sequence of \textit{subwords}. We prepend a space to every first subword in a word and use no special marker for other subwords. Note that the subwords might not correspond directly to parts of input, because the \textit{bert uncased} model strip input casing and diacritics.


\subsection{Alignment}

\setlength{\algomargin}{0em}
\SetAlCapHSkip{0em}
\begin{algorithm}[t]
  \DontPrintSemicolon

  \SetKwInput{Input}{Input}
  \SetKw{Continue}{continue}
  
  \Input{input subwords $\bm{s}$, gold characters $\bm{g}$}
  $\bm{w}[:, :] \gets \bm{0}$\;
  \For{$i\gets |\bm{s}|$ \KwTo $1$}{
    \For{$j\gets |\bm{g}|$ \KwTo $1$}{
      $\bm{w}[i, j] \gets \bm{w}[i+1, j]$\;
      \For{$l \gets 1$ \KwTo $\max(|\bm{g}| - j, 8 + 3|\bm{s}[i]|)$}{
        $g \gets \bm{g}[j:j+l]$\;
        \lIf{$g$.isspace()}{\Continue}
        $c \gets \begin{cases}
          1\hphantom{.75} \textbf{~~if } \bm{s}[i]=g \\
          0.75 \textbf{~~if } \bm{s}[i]\textit{.strip()}=g\rlap{\textit{.strip()}}\\
          0.5 \cdot \textrm{LevenshSimilarity}\rlap{$(\bm{s}[i], g)$}
        \end{cases}$\;
        $\bm{w}[i, j] \gets \max(\bm{w}[i, j], c + \bm{w}[i, j+1])$
      }
    }
  }
  \Return{\rm alignment with weight $\bm{w}[0,0]$, as in LCS}
  
  \caption{Extended LCS Alignment}\label{alg:lcs}
\end{algorithm}

\looseness-1
In order to \textit{automatically} encode the gold data via \textit{character} transformations, we first align the input subwords and the corrected sentence. We compute the alignment with Algorithm~\ref{alg:lcs}, which is an extended version of LCS, where each subword is aligned with a \textit{sequence} of gold characters. We ignore casing, diacritics and consider all punctuation equal during the alignment, and bound the maximum length of a correction (number of characters aligned to a single input subword) by $8+3 \cdot \textit{input subword character length}$ for efficiency\rlap{.}

%

\subsection{Transformations}

\begin{figure}[t]
    \centering
    \small
    \setlength{\tabcolsep}{2pt}
    \begin{tabular}{@{}lllll@{}}
    \toprule
      \textbf{Input} & \multicolumn{2}{l}{gatherin} & \multicolumn{2}{l}{leafes} \\[2pt]
      \textbf{Correct} & \multicolumn{2}{l}{\textbf{G}atherin\textbf{g}} & \multicolumn{2}{l}{lea\textbf{v}es} \\
    \midrule  
      \textbf{Subwords} & \textvisiblespace gathe & rin & \textvisiblespace lea & fes \\
    \midrule  
      \textbf{string-at-word} & \textvisiblespace Gathering &  & \textvisiblespace leaves &  \\[3pt]
      \textbf{string-at-subword} & \textvisiblespace Gathe & \textsc{app}. g & \textsc{keep} & ves \\[2pt]
      \textbf{char-at-word} & \makecell[l]{\textsc{append} g,\\\textsc{upperc}. $2^\textrm{nd}$} &  & \makecell[l]{\textsc{repl}. \rlap{$3^\textrm{rd}$}\\\textit{from \rlap{end}}\\\textit{with} v} & \\[2pt]
      \textbf{char-at-subword} & \textsc{upperc}. $2^\textrm{nd}$ & \textsc{app}. g & \textsc{keep} & \makecell[l]{\textsc{repl}. $1^\textrm{st}$\\\textit{with} v} \\
    \bottomrule
    \end{tabular}
    \caption{Example of the four types of transformations.}
    \label{fig:example}
\end{figure}


We consider four kinds of transformations, differing in two dimensions -- the granularity of the transformation and the unit it is applied on:

\begin{citemize}
  \item character transformations applied on each subword separately (\emph{char-at-subword}),
  \item character transformations applied on each complete word (\emph{char-at-word}),
  \item string transformations applied on each subword (\textit{string-at-subword}),
  \item string transformations applied on each complete word (\textit{string-at-word}).
\end{citemize}

\noindent In such terminology, the transformations proposed by \citet{awasthi-etal-2019-parallel} and \citet{omelianchuk-etal-2020-gector} can be referred to as \textit{string-at-word}. An example of the described transformation types is illustrated in Figure~\ref{fig:example}.


To apply a transformation on a complete word, we concatenate the corresponding subwords and aim to produce the concatenation of the subwords' corrections.

A string transformation can be one of \textit{keep}, \textit{replace} by given string or \textit{append} a given string before/after.


A character transformation consists of multiple character edits, which we construct as follows:
\begin{cenumerate}
\item We start by computing the standard smallest \textit{edit script} between an input subword and a correction. The edit script is a sequence of \textit{inserts}, \textit{replaces} and \textit{deletes}, and we index each edit operation either from the beginning of the input subword (if it involves the first half of it) or from the end of it (otherwise). The edit script is computed on lowercased strings, and in case of \textit{bert uncased} models, also on undiacritized strings.
\item Afterwards, the unmodified input subword (i.e., including casing in case of \textit{bert cased} models) is processed by the edit script, obtaining a correction with possibly incorrect casing. If some incorrectly lowercased characters are indeed present, \textit{uppercase} operations are added, each indexing a single character either from the beginning of the correction (if the character is in the first half of it) or from the end of it (otherwise).
\item Finally, for \textit{bert uncased} models, we still need to handle missing diacritics. We achieve it analogously to step 2, adding missing diacritical marks with operations indexing single characters again either from the beginning of the correction (if the character is in the first half) or from the end of it.
\end{cenumerate}

\noindent The reason for special handling of casing (and diacritization for \textit{bert uncased} models) is that the proposed rules are more general, allowing to capture for example corrections \textit{go$\rightarrow$Going} and \textit{walk$\rightarrow$Walking} with a single rule \textit{append ``-ing'', uppercase first}.

  


\section{Transformations Upper-bound F-score}
\label{sec:coverage}

\begin{figure*}[t!]
  \includegraphics[width=\hsize]{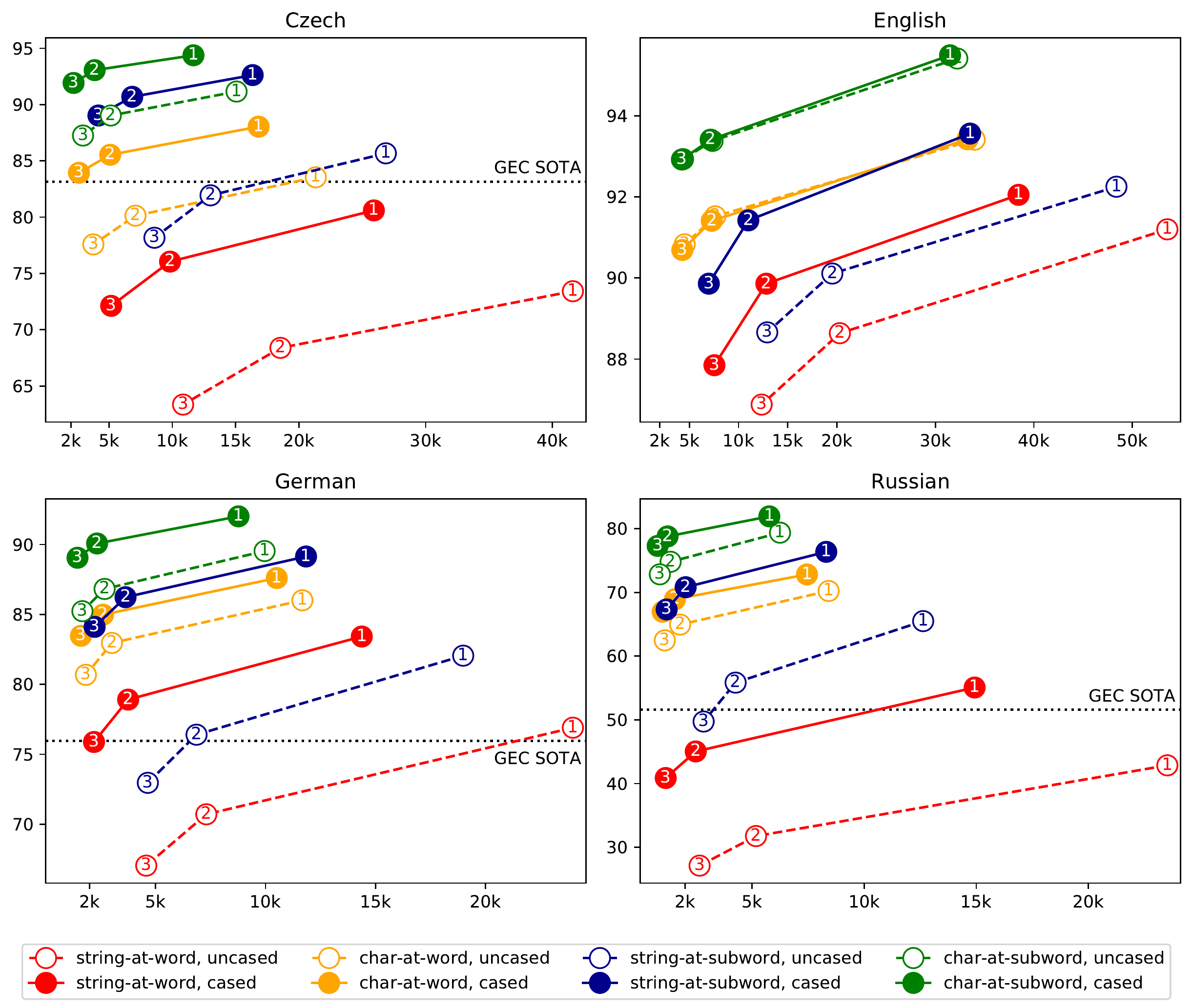}
  \caption{F$_{0.5}$ depending on number and type of transformations, if all transformations were correctly predicted (upper-bound). Up and right is better (higher F$_{0.5}$, fewer rules), down and left is worse (lower F$_{0.5}$, more rules). Circled numbers \textcircled{1}, \textcircled{2} and \textcircled{3} denote that we kept transformations present at least once, twice or three times in the training data, respectively (larger means less transformations).}
  \label{fig:coverage}
\end{figure*}

To assess the effect of number and type of transformations, we compute the potential maximum $F_{0.5}$ score with the MaxMatch M$^2$ scorer \cite{dahlmeier-ng-2012-better}.


We generate transformation dictionary from the training portion of the following GEC corpora: CoNLL-2014 shared task for English \cite{ng-etal-2014-conll}, AKCES-GEC \cite{naplava2019grammatical} for Czech, FALKO-MERLIN GEC \cite{boyd-2018-using} for German and RULEC-GEC \cite{rozovskaya2019grammar} for Russian. The sizes of the datasets are quantified in Table~\ref{tab:data_overview}.

We also generate transformations from synthetic data augmentation used for training (Section~\ref{sec:experiments}); to prevent the explosion of the transformation dictionary, we consider only 1000 synthetic sentences, except for Russian, which employs 5000 synthetic sentences because of very small authentic data. Finally, we add a special \textit{uncorrectable error} dictionary element, indicating an error that cannot be corrected by any dictionary transformation. 

To encode a gold correction with a transformation, we first try looking it up in the dictionary. If it is not present, we go through all dictionary transformations in random order, accepting the first one producing the correct output. If no transformation match, we resort to the \textit{uncorrectable error} (during prediction, it will be replaced by the input token).

We show all combinations (\textit{character/string} at \textit{words/subwords}), using cased and uncased mBERT, in Figure~\ref{fig:coverage}. Clearly, character transformations applied at subwords (\textit{char-at-subword}, green) have the highest potential in terms of upper-bound F$_{0.5}$ in all four languages. At the same time, word replacements (\textit{string-at-word}, red) do not scale well. This effect is emphasized in morphologically-rich Czech and Russian, for which the upper-bound string replacement F$_{0.5}$ (\textit{string-at-word}, red) falls below the current GEC systems state-of-the-art F$_{0.5}$ (horizontal dotted line).

\begin{table}[t]
\centering
\begin{tabular}{llc}
\toprule
Language & Dataset & Sentences \\
\midrule
 Czech      & AKCES-GEC         & 47\,371                    \\
 German     &  Falko-MERLIN & 24\,077 \\
 Russian & RULEC-GEC & 12\,480 \\
 \bottomrule
\end{tabular}
 \caption{GEC datasets used for constructing rules and for evaluation, including their size.}
    \label{tab:data_overview}
\end{table}

\section{Experiments}
\label{sec:experiments}


We train the character subword GEC tagging model using the \textit{char-at-subword} transformation, which have achieved the best upper-bound score.\footnote{We also performed preliminary experiments with \textit{char-at-word} GEC tagging model, and it performed worse than using the \textit{char-at-subword} transformations.} We train the character subword GEC tagging model (\textit{char-at-subword}) for Czech, German and Russian, in two stages: First, models are trained on a large synthetic corpus, generated by a reimplementation of \citet{naplava2019grammatical}. Then, the models are finetuned on a mixture of synthetic and authentic data in ratio 1:2. The authentic data used in the second stage are AKCES-GEC~\cite{naplava2019grammatical} for Czech, FALKO-MERLIN GEC~\cite{boyd-2018-using} for German and RULEC-GEC~\cite{rozovskaya2019grammar} for Russian.

The model is based on a pretrained BERT encoder \cite{devlin-etal-2019-bert}, specifically \textit{bert-base-multilingual} (\textit{uncased} for Czech and German, \textit{cased} for Russian). After encoding the tokens, we add a simple softmax classifier that projects embeddings for each subword into a distribution over a set of transformations (Section~\ref{sec:transformations}) generated from authentic data with a limited addition from synthetic data (Section~\ref{sec:coverage}). We generate 7.7k transformations for Czech, 4.3k transformations for German and 3.1k transformations for Russian.

GEC models based on Transformer and BERT-encoder were shown to perform better when applied iteratively~\cite{lichtarge2018weakly, omelianchuk-etal-2020-gector}. Therefore, we experiment with multiple iterations and report results both for single iteration and four iteration phases after which we did not observe significant improvements.

We train both the fully-connected network and BERT with AdamW optimizer~\cite{loshchilov2018decoupled} which minimizes the negative log-likelihood. Both for pretraining and finetuning, the learning rate linearly increases from 0 to $5 \cdot 10^{-5}$ over the first 10000 steps and linearly decreases to 0 over 20 epochs. We use the batch size of 2048 sentences and clip each training sentence to 128 tokens. We pretrain each model for circa 14 days and finetune it for circa 2 days on Nvidia P5000 GPU and select the best checkpoint according to development set.

We experimented with weighting all classes different from the KEEP instruction by a factor of 3. It turned out effective only for pretraining Russian.

\begin{table}[t!]
    \centering
    \setlength{\tabcolsep}{5.75pt}
    \begin{tabular}{lcc}
        \toprule
        Model & \llap{P}arams & F$_{0.5}$ \\
        \midrule
        \citet{richter2012korektor} & & 58.54 \\
        \citet{naplava2019grammatical}$^\mathit{synt}$ & 210M & 66.59 \\
        \citet{naplava2019grammatical}$^\mathit{fine}$ & 210M & 80.17 \\
        \citet{Rothe-2021-mT5-GEC-SOTA} \textit{base} & 580M & 71.88 \\
        \citet{Rothe-2021-mT5-GEC-SOTA} \textit{xxl} & 13B & 83.15 \\
        \midrule
        Ours synthetic & 172M & 64.29 \\ 
        Ours finetuned & 172M & 72.86 \\
        Ours finetuned 4 iterations & 172M & 75.06 \\
        \bottomrule
    \end{tabular}

    \medskip
    \centerline{(a) Czech}
    \bigskip
    
    \begin{tabular}{lcc}
        \toprule
         Model & \llap{P}arams & F$_{0.5}$  \\
        \midrule 
         \citet{boyd-2018-using} & &  45.22 \\
        \citet{naplava2019grammatical}$^\mathit{synt}$ & 210M & 51.41 \\
        \citet{naplava2019grammatical}$^\mathit{fine}$ & 210M & 73.71 \\
        \citet{Rothe-2021-mT5-GEC-SOTA} \textit{base} & 580M & 69.21 \\
        \citet{Rothe-2021-mT5-GEC-SOTA} \textit{xxl} & 13B & 75.96 \\
        \midrule
        Ours synthetic & 170M & 44.29 \\
        Ours finetuned & 170M & 62.92 \\
        Ours finetuned 4 iterations & 170M & 65.95 \\
        \bottomrule
    \end{tabular}
    
    \medskip
    \centerline{(b) German}
    \bigskip
    

    \begin{tabular}{lcc}
        \toprule
         Model & \llap{P}arams & F$_{0.5}$  \\
        \midrule
        \citet{rozovskaya2019grammar} & & 21.00 \\
        \citet{naplava2019grammatical}$^\mathit{synt}$ & 210M & 40.96 \\
        \citet{naplava2019grammatical}$^\mathit{fine}$ & 210M & 50.20 \\
        \citet{Rothe-2021-mT5-GEC-SOTA} \textit{base} & 580M & 26.24 \\
        \citet{Rothe-2021-mT5-GEC-SOTA} \textit{xxl} & 13B & 51.62 \\
        \midrule
        Ours synthetic & 180M & 25.36 \\
        Ours finetuned & 180M & 36.62 \\
        Ours finetuned 4 iterations & 180M & 38.68 \\
        \bottomrule
    \end{tabular}
    
    \medskip
    \centerline{(c) Russian}
    \bigskip
    
    \caption{Model results}
    \label{tab:results}
\end{table}

We present the results of our models in Table~\ref{tab:results}. Compared to autoregressive models of similar size \cite{naplava2019grammatical}, our models achieve solid results with large speedup due to the non-autoregressive tagging approach. Obviously, the inflation of model size \cite{Rothe-2021-mT5-GEC-SOTA} to enormous size (13B parameters) leads to further improvements at the cost of increased computational demands.

\subsection{Runtime Performance}

\begin{table}[t!]
    \centering
    \setlength{\tabcolsep}{5.75pt}
    \begin{tabular}{lc}
        \toprule
        Model & Time Per Sentence \\
        \midrule
        T2T & 162.34 \\
        BERT-GEC & ~~41.26 \\
        \bottomrule
    \end{tabular}

    \medskip
    \centerline{(a) CPU decoding on a 32-core Intel Xeon}
    \bigskip
    
    \begin{tabular}{lc}
        \toprule
         Model & Time Per Sentence \\
        \midrule 
        T2T & 22.36 \\
        BERT-GEC & ~~5.09 \\
        \bottomrule
    \end{tabular}
    
    \medskip
    \centerline{(b) GPU decoding on Nvidia Quadro P5000}
    \medskip

    \caption{Average time in milliseconds required to process a single sentence in the Czech test set, measured using both (a) CPU decoding and (b) GPU decoding.}
    \label{tab:speedup}
\end{table}

To evaluate the speed-up of the non-autoregressive decoding, we compare the runtime performance of our system to the autoregressive Transformer encoder-decoder architecture from \citet{naplava2019grammatical}, which is of comparable size. The measurements are performed using both a CPU-only decoding (performed on a dedicated 32-core Intel Xeon E5-2630) and GPU decoding (measured on an Nvidia Quadro P5000). The results presented in Table~\ref{tab:speedup} show that the non-autoregressive system is four times faster.



\section{Conclusion And Future Work}

We proposed a character-based method to generate target transformation instructions for GEC tagging models, as an alternative to autoregressive models. We compared the character transformations to previously used word-level transformation instructions and have shown that character-based rules have better coverage and scale better in Czech, German and Russian. Moreover, we trained character-based GEC tagging models for these languages. The source code is available at {\small\url{https://github.com/ufal/wnut2021_character_transformations_gec}}.

For future work, we propose to investigate ways to generate synthetic data to achieve better coverage of the target transformation set, since the current process for generating synthetic errors is well suited for encoder-decoder models, but may fail to cover certain transformations.

\section*{Acknowledgements}

The research described herein has been supported by the Ministry of Education, Youths and Sports of the Czech Republic, under the project LINDAT/CLARIAH-CZ (LM2018101).
This research was also partially supported by SVV project number 260 575 and GAUK 578218 of the Charles University.

\bibliography{anthology,custom}
\bibliographystyle{acl_natbib}

%
%

\end{document}